\documentclass[runningheads]{llncs}
\usepackage{amsmath,amssymb}
\usepackage[square,numbers]{natbib}
\usepackage{booktabs}
\usepackage{tabularx}
\usepackage{multirow}
\usepackage{graphicx}
\usepackage{todonotes}
\usepackage{longtable}
\usepackage[normalem]{ulem}
\usepackage{siunitx}
\usepackage{colortbl}

\useunder{\uline}{\ul}{}

\definecolor{gray1}{HTML}{F5F5F5}
\definecolor{gray2}{HTML}{E5E5E5}
\definecolor{gray3}{HTML}{D5D5D5}
\definecolor{gray4}{HTML}{C5C5C5}
\definecolor{gray5}{HTML}{B5B5B5}
\definecolor{gray6}{HTML}{A5A5A5}
\definecolor{gray7}{HTML}{959595}
\definecolor{gray8}{HTML}{858585}

\begin{document}

\title{Zero-shot performance of the Segment Anything Model (SAM) in 2D medical imaging: A comprehensive evaluation and practical guidelines}
\titlerunning{Zero-Shot SAM in 2D Medical Imaging: Evaluation \& Guidelines}

\author{Christian Mattjie\inst{1} \and
Luis Vinicius de Moura\inst{1} \and
Rafaela Cappelari Ravazio\inst{1} \and 
Lucas Silveira Kupssinskü\inst{1}\and 
Otávio Parraga\inst{1}\and 
Marcelo Mussi Delucis\inst{1}\and 
Rodrigo C. Barros\inst{1}}
\authorrunning{Christian Mattjie et al.}

\institute{
    School of Technology, Pontif{\'i}cia Universidade Cat{\'o}lica do Rio Grande do Sul \\
    Av. Ipiranga, 6681, 90619-900, Porto Alegre, RS, Brazil \\
    \email{christian.oliveira95@edu.pucrs.br} \\
    \email{rodrigo.barros@pucrs.br}
}

\maketitle
\begin{abstract}
Segmentation in medical imaging is a critical component for the diagnosis, monitoring, and treatment of various diseases and medical conditions. 
Presently, the medical segmentation landscape is dominated by numerous specialized deep learning models, each fine-tuned for specific segmentation tasks and image modalities. 
The recently-introduced Segment Anything Model (SAM) employs the ViT neural architecture and harnesses a massive training dataset to segment nearly any object; however, its suitability to the medical domain has not yet been investigated. 
In this study, we explore the zero-shot performance of SAM in medical imaging by implementing eight distinct prompt strategies across six datasets from four imaging modalities, including X-ray, ultrasound, dermatoscopy, and colonoscopy. 
Our findings reveal that SAM's zero-shot performance is not only comparable to, but in certain cases, surpasses the current state-of-the-art. Based on these results, we propose practical guidelines that require minimal interaction while consistently yielding robust outcomes across all assessed contexts. 
The source code, along with a demonstration of the recommended guidelines, can be accessed at \url{https://github.com/Malta-Lab/SAM-zero-shot-in-Medical-Imaging}.

\end{abstract}
\section{Introduction}

Medical imaging plays a pivotal role in the diagnosis, monitoring, and treatment of a wide range of diseases and conditions~\citep{wang2017central}. 
Accurate segmentation of these images is often critical in extracting valuable information that can aid clinical decision-making. 
However, traditional segmentation methods primarily rely on labor-intensive, manually-engineered features and error-prone thresholding designed for specific scenarios, resulting in limited generalizability to new images~\citep{pham2000current}. 
Large advancements in medical image segmentation have been achieved with the advent of deep learning (DL) techniques, owing to their ability to learn intrinsic features and patterns from large datasets~\citep{unet, universal, yang2023rema}.

The DL revolution was ignited by the groundbreaking success of Convolutional Neural Networks (CNNs) in computer vision applications~\citep{litjens2017survey}. 
Recently, a new wave of innovative applications based on the Transformer architecture has emerged~\citep{dosovitskiy2020image}. 
Transformers enhance the training process by harnessing larger datasets while providing smaller induction bias, thereby creating models that can generalize to unseen distributions and even adapt to diverse tasks.

Nonetheless, medical image segmentation poses significant challenges for DL due to the substantial cost associated with specialized professionals annotating images, leading to the scarcity of available data. 
Furthermore, there is limited evidence regarding the ability of DL models trained on natural images to generalize to medical application settings.

The Segment Anything Model (SAM) has been recently introduced by Meta~\citep{SAM}. 
SAM, a state-of-the-art vision transformer (ViT), is capable of generating segmentation masks for virtually any object. 
It introduces the concept of prompting in image segmentation, whereby the model's inference process is guided by providing points inside the region of interest (ROI) or by drawing a bounding box around it.

In this paper, we rigorously evaluate the zero-shot capabilities of SAM in segmenting 2D medical images. 
We assess its performance across six datasets encompassing four distinct imaging modalities: X-ray, ultrasound, dermatoscopy, and colonoscopy, using various prompting strategies. 
Our comprehensive evaluation reveals that SAM demonstrates promising results in those medical imaging modalities, even when we have complex patterns such as hair on skin lesions. We also propose practical guidelines for physicians to utilize SAM in medical image segmentation tasks. This guideline suggests starting with a bounding box prompt, selecting the optimal prediction from the generated outputs, and refining the segmentation using point prompts when necessary.
\section{Related Work}

\subsection{Medical Image Segmentation}

Medical image segmentation plays a pivotal role in medical imaging analysis, focusing on the identification and delineation of structures or regions such as organs, tissues, or lesions. Accurate segmentation is crucial for various clinical applications, encompassing diagnosis, treatment, and monitoring of disease progression. This enables essential tasks like measuring tissue volume for tracking growth and outlining radiosensitive organs in radiotherapy treatment.

In the current domain of medical image segmentation, specific methods are tailored to the application, imaging modality, and body part under examination~\citep{chowdhary2020segmentation,sharma2010automated}. However, automatic segmentation remains a formidable challenge due to the intricacy of medical images and data scarcity. The segmentation algorithm's output is influenced by multiple factors, including the partial volume effect, intensity inhomogeneity, presence of artifacts, and insufficient contrast between soft regions~\citep{shirly2019review}. 

Deep learning techniques have garnered considerable attention in medical image segmentation, owing to their capacity for capturing intricate patterns and representations from large-scale datasets. Among the most prevalent DL approaches for medical image segmentation are CNNs. Widely employed models for medical image segmentation include U-Net~\citep{unet} and its derivatives, which were explicitly developed for biomedical image segmentation. U-Net utilizes a symmetric encoder-decoder architecture, enabling the model to capture both high-level contextual information and fine-grained details, resulting in enhanced segmentation outcomes.

In recent years, novel state-of-the-art segmentation techniques have emerged, such as training DL models on polar images~\citep{polar}, integrating textual information with vision-language models~\citep{universal}, and employing attention mechanisms with CNNs in ViTs~\citep{henry2022vision}.

\subsection{Vision Transformer (ViT)}

ViTs constitute a class of DL models that leverage the transformer architecture~\citep{vaswani2017attention}. These models process images by dividing them into fixed-size, non-overlapping patches and linearly embedding these patches into a flat sequence of tokens. Each token is subsequently passed through a series of self-attention layers to learn relevant contextual relationships and spatial information, enabling the model to discern semantically-rich patterns~\citep{dosovitskiy2020image}.

ViTs do not share some of the inductive biases inherent in CNNs, such as locality and translation equivariance. A reduced inductive bias allows ViTs to be more adaptable even though it necessitates more data for generalization. The data demand may limit the application of ViTs in medical imaging, where data is often scarce. Nevertheless, by capitalizing on pre-training and fine-tuning strategies, ViTs are revolutionizing the computer vision landscape with strong generalization performance~\citep{touvron2021training,wang2021pyramid}.

Recently, ViTs have demonstrated strong results in zero-shot learning~\citep{guo2022calip,pham2021combined,radford2021learning}. This setting presents a challenge since the model must learn to generalize for classes and contexts not encountered during training. In medical imaging, ViT-based models have achieved state-of-the-art results~\citep{jang2022significantly, wang2022medclip, henry2022vision}, though very few studies address the zero-shot capabilities of the learning models, and whether their performance in zero-shot settings is reasonable or even competitive to fine-tuning \citep{guo2022zero, henry2022vision}.
\section{Methodology}

\subsection{Segment Anything Model (SAM)}

SAM~\citep{SAM} is a state-of-the-art ViT model trained on the massive SA-1B dataset (also introduced in \citep{SAM}). 
This dataset comprises approximately 11 million images and 1 billion segmentation masks, making it the largest publicly available image segmentation dataset to date. 
The model's high accuracy has been demonstrated through its impressive capability of segmenting a wide variety of objects and shapes, thereby validating its effectiveness in segmenting virtually any object within a 2D image.

SAM can function in two distinct ways: by segmenting all objects present in an input image or by utilizing prompts that explicitly specify the target region for segmentation. These prompts can take the form of points identifying the region of interest or regions that should be excluded. Additionally, a bounding box may be provided to delineate the area containing the object of interest. While initial results with SAM showcase strong segmentation quality and zero-shot generalization to novel scenes and unseen objects, it is important to note that the model's training dataset lacks medical images. Consequently, its generalizability to the medical domain remains an open question.

To address potential issues arising from ambiguous prompts, SAM generates a set of three masks, each with an accompanying score reflecting a different interpretation of the intended region. The first mask in the output sequence represents the smallest, most conservative interpretation of the intended region according to the given prompt. As the sequence progresses, the subsequent masks increase in size, with each mask encompassing the previous one. The score assigned to each mask is an indicator of SAM's confidence in that particular prediction. This design enables SAM to accommodate a wider range of potential segmentation outcomes, reflecting the model's efforts to account for the ambiguity in the target region's size due to the prompt's limited information.

In practical applications, especially within the medical imaging domain, it is crucial to ensure that the model accurately identifies and segments pertinent structures or regions of interest. Given this requirement, our study focused on investigating input prompt strategies for guiding SAM's segmentation process. This decision stems from the inherent uncertainties associated with the segment-everything approach, as the model's comprehension of the segmented objects cannot be guaranteed. By utilizing prompts, we aimed to improve SAM's segmentation capabilities in medical imaging tasks and provide a more reliable and controlled evaluation of its performance. Furthermore, we did not consider the confidence scores provided by SAM for each mask, as these scores reflect the quality of the segmentation without accounting for the accuracy of the target region relative to the intended object.

The ViT architecture employed by SAM consists of three distinct iterations, each with unique trade-offs between computational requirements and model performance: ViT Base (ViT-B), ViT Large (ViT-L), and ViT Huge (ViT-H). The primary differences between these iterations lie in the model's number of layers and parameters, as illustrated in Table~\ref{tab:sam}. As the number of layers and parameters increases, the model becomes more powerful, enabling the capture of more intricate aspects of the input images. However, larger models necessitate more computing power, which may pose a drawback in certain situations. Nevertheless, even the largest iteration of SAM remains relatively compact.

\begin{table}[!htpb]
    \centering
    \begin{tabularx}{\textwidth}{@{}>{\centering\arraybackslash}X>{\centering\arraybackslash}X>{\centering\arraybackslash}X>{\centering\arraybackslash}X@{}}
        \toprule
        Architecture & Transformer Layers & Parameters & Size (Mb) \\
        \midrule
        ViT-B & 12 & 91M & 776 \\
        ViT-L & 24 & 308M & 1582 \\
        ViT-H & 32 & 636M & 2950 \\
        \bottomrule
    \end{tabularx}
    \vspace{0.7mm}
    \caption{Summary of SAM's ViT architecture variations.}
    \vspace{-5mm}
    \label{tab:sam}
\end{table}

\subsection{Datasets}

\begin{figure}[!htpb]
    \centering
    \includegraphics[width=.80\textwidth]{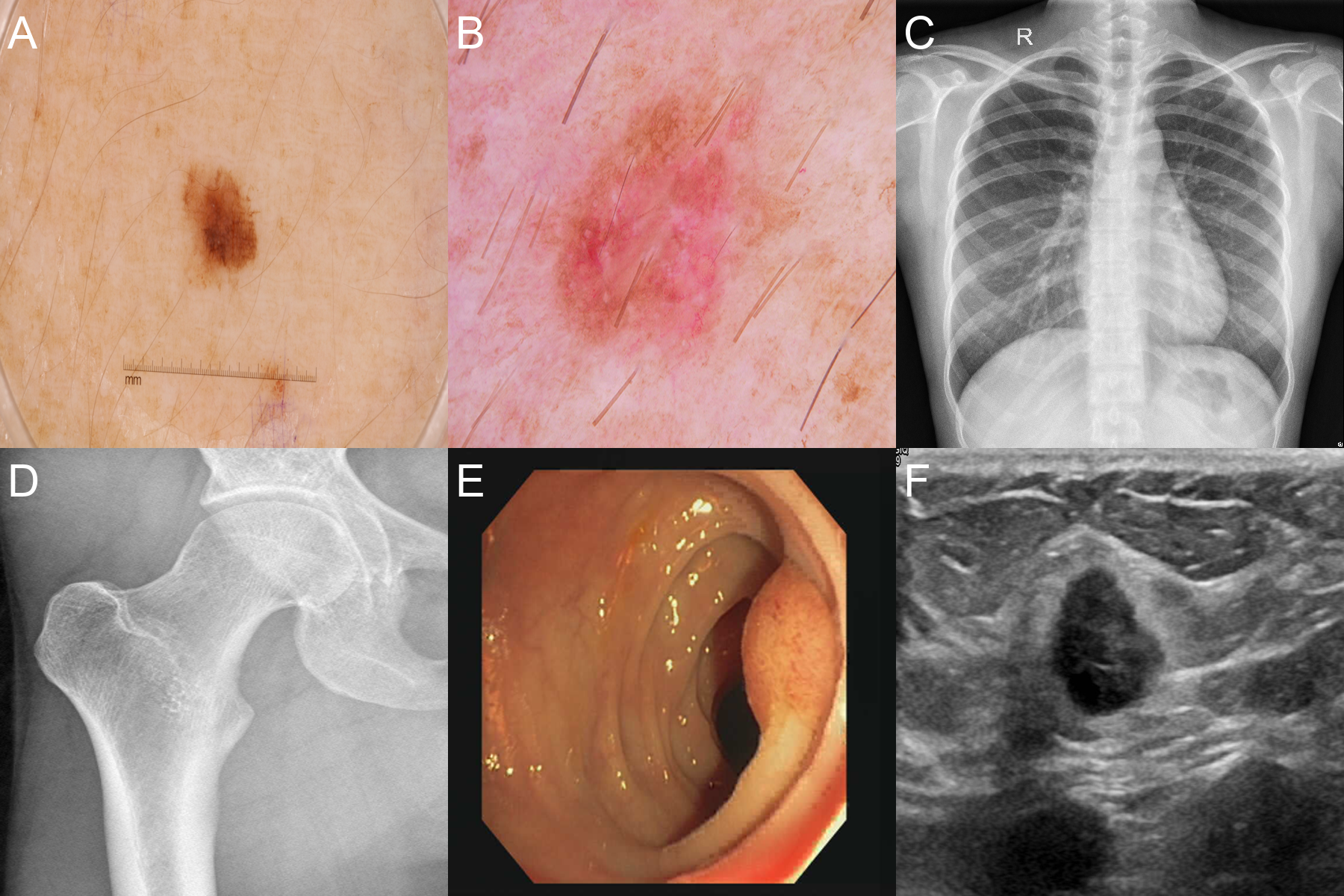}
    \caption{Samples from each of the six datasets used in this study. A: ISIC, B: HAM, C: CXR, D: HJXR, E: CVC, F: BUSI.}
    \label{fig:datasets}
\end{figure}

For evaluating SAM, we used six datasets from four medical imaging modalities: X-ray, Ultrasound, dermatoscopic, and colonoscopy images. Our primary objective is to assess the model's performance and versatility when prompted with various strategies, simulating a physician's approach to segmenting specific organs or ROIs in medical images. Fig~\ref{fig:datasets} shows a sample from each dataset.

\begin{itemize}
    \item \textbf{ISIC 2018}~\citep{ISIC2018}: this publicly available dataset comprises $2,594$ dermatoscopic images from $2,056$ unique patients, showcasing skin lesions with varying types, sizes, and colors. The images have resolutions ranging from $640\times 480$ to approximately $6,700\times 4,400$ pixels and are provided in JPEG format. Expert dermatologists generated accompanying segmentation masks using a manual annotation tool, and a second expert reviewed each mask for accuracy.
    
    \item \textbf{HAM10000}~\citep{HAM}: this dataset contains $10,015$ dermatoscopic images of skin lesions from $7,388$ unique patients, with varying types, sizes, and colors. All images have a resolution of $640\times 450$ and are provided in JPEG format. Recently, Tschandl, P. et al.\citep{HAMseg} supplied expert segmentation masks for all images, with corresponding resolutions.
    
    \item \textbf{Montgomery-Shenzhen}~\citep{CXR1, CXR2}: this dataset is a fusion of two publicly available chest X-ray datasets collected from respective hospitals. It comprises $800$ X-ray images, with $704$ accompanying lung segmentation masks manually created by expert radiologists. The dataset is available in PNG format.
    
    \item \textbf{X-ray Images of Hip Joints}~\citep{hipjoints}: this publicly available dataset contains $140$ X-ray images of the lower legs, with an average resolution of $327\times 512$. Corresponding segmentation masks for the femur and ilium are provided separately. The images and masks are available in NII format.
    
    \item \textbf{CVC-ClinicDB}~\citep{CVC}: this dataset consists of $612$ images from $31$ colonoscopy sequences, with a resolution of $384\times 288$. The images are provided in PNG format. Expert gastroenterologists have created segmentation masks for the polyps, which are provided for all available images.
    
    \item \textbf{Breast Ultrasound Images}~\citep{BreastUS}: this dataset comprises $780$ ultrasound images of the breast from $600$ patients, with an average size of $500\times 500$ pixels. The images are provided in PNG format and are categorized into normal, benign, and malignant. Segmentation masks for tumors are supplied for both benign and malignant cases.
\end{itemize}

\subsection{Prompt Strategies}
\label{prompt_strat}

In the context of interactive segmentation, a physician may guide the procedure using various strategies, such as clicking within the region of interest, clicking outside the region, or drawing a bounding box around the target. To investigate the impact of these plausible prompting strategies on our segmentation models, we conducted a series of experiments with the following approaches:

\begin{itemize}
    \item \textbf{Central-point (CP)}: utilizing only the centroid of the ground-truth mask, which is anticipated to be the most informative single-point prompt;
    \item \textbf{Random-point (RP)}: eroding the ground-truth mask and subsequently selecting a random point within it, representing minimal guidance;
    \item \textbf{Distributed random-points (RP3 and RP5)}: eroding the ground-truth mask, dividing it vertically into sections (three and five, respectively), and selecting a random point within each section to provide a more distributed set of prompts;
    \item \textbf{Bounding-box (BB)}: prompting with the bounding box of the ground-truth mask, offering a more explicit spatial constraint for segmentation; and
    \item \textbf{Perturbed bounding-box (BBS5, BBS10, and BBS20)}: modifying the size and position of the bounding box by $5$\%, $10$\%, and $20$\% of the ground-truth mask size, respectively, simulating variations in the accuracy of a physician's initial assessment.
\end{itemize}

\begin{figure}[!hptb]
    \centering
    \includegraphics[width=.80\textwidth]{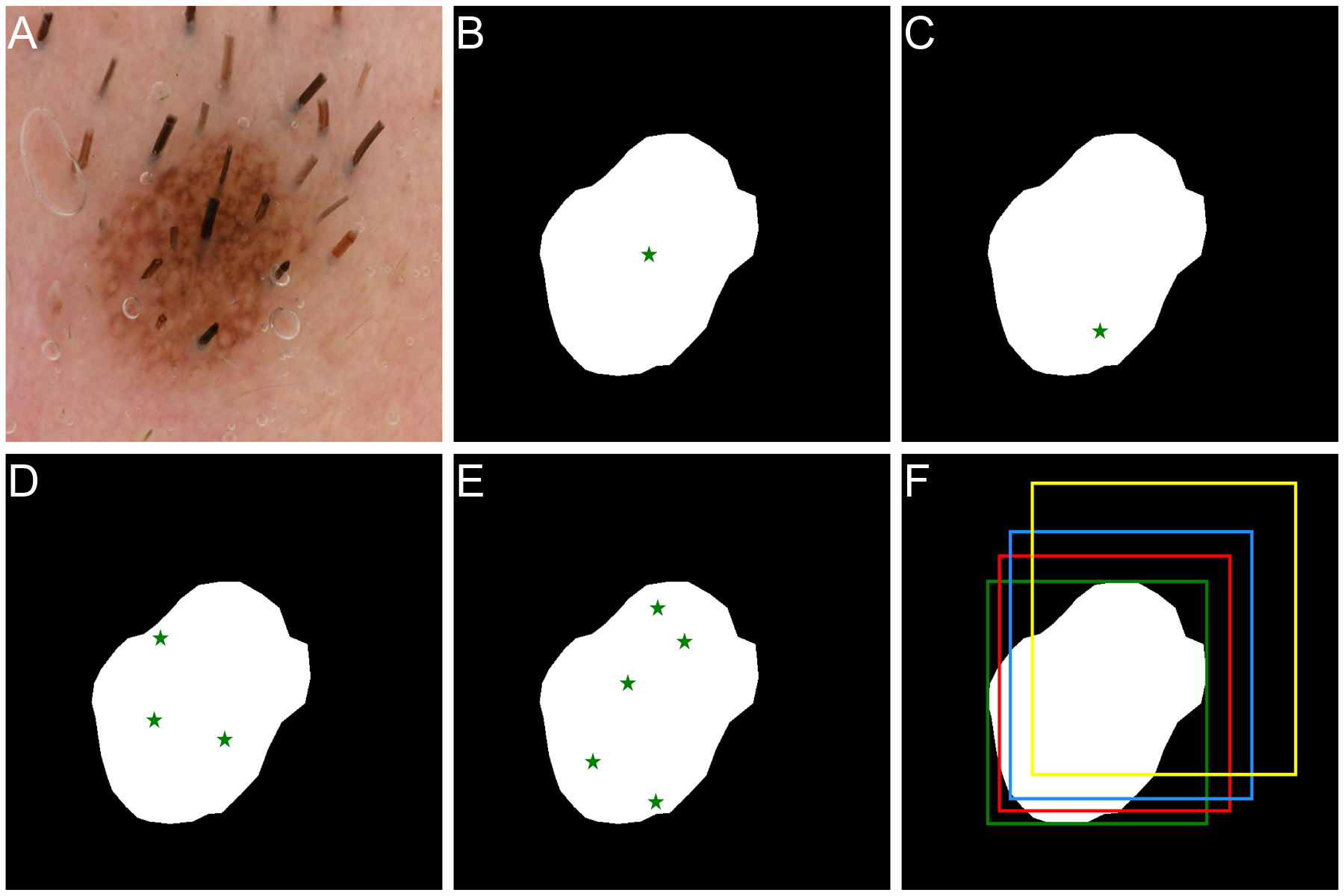}
    \caption{Example of all prompt strategies on a skin lesion image and mask. (A): original image, (B): CP, (C): RP, (D): RP3, (E): RP5, (F): BB in green, BBS5 in red, BBS10 in blue, and BBS20 in yellow. The size and position shown are their max variation for BBS methods.}
    \label{prompts}
\end{figure}

For the multiple points strategy, we divided the mask into three and five sections, and for the varied bounding box strategy, we randomly altered its size and position up to 5\%, 10\%, and 20\% of the ground-truth mask. 
Given these variations, we ran a total of eight experiments per model/dataset, which are shown in Fig~\ref{prompts}: central-point (CP), random-point (RP), random-points-3 (RP3), random-points-5 (RP5), bounding-box (BB), bounding-box-similar-5 (BBS5), bounding-box-similar-10 (BBS10) and bounding-box-similar-20 (BBS20).

In the RP, RP3, and RP5 strategies, we apply an erosion morphological operator to the ground-truth mask before selecting a random point within the resulting region. This process ensures that the selected point is not situated near the region of interest's edges while preserving the element of randomness expected in real-world scenarios. The erosion value was determined according to the dataset: for the CXR and ISIC datasets, where the regions are larger, we used a $30$-pixel radius; for the CVC dataset, which contains smaller images that could be completely eroded, we employed a $1$-pixel radius; for all other datasets with relatively small regions, we opted for a $10$-pixel radius.

The various prompting strategies applied to a skin lesion image and mask are illustrated in Fig~\ref{prompts}. This figure demonstrates the original image (A), and the different prompt strategies: CP (B), RP (C), RP3 (D), RP5 (E), and BB (F) in green, BBS5 in red, BBS10 in blue, and BBS20 in yellow. The size and position shown for the BBS methods represent their maximum variation, while in our experiments, they were altered randomly.

\subsection{Preprocessing}

Throughout our experimentation process, we encountered various challenges stemming from the characteristics of the datasets under consideration. To address these issues, we implemented a fill-holes technique aimed at rectifying mask information, particularly in the context of the ISIC dataset, wherein some masks solely outlined the relevant lesion. Moreover, in instances where an image contained multiple masks (e.g., dual lungs or skin lesions), we isolated the two most substantial regions and processed them independently, employing the prompt strategies delineated in the previous section. This approach ensured accurate and precise segmentation. A manual inspection of all images was conducted to confirm the absence of any containing three distinct and relevant regions. The model-generated predictions for both regions were combined to form a single prediction. 

In the case of the HJXR dataset, which uniquely contained images in a format incompatible with SAM, we transformed the images from NII to PNG format, normalizing their values within the range of $0$ to $255$. Given that masks for the femur and ilios were available individually for each image in the dataset, we assessed the predictions separately in HJXR-F and HJXR-I.

\subsection{Evaluation}

The Dice Similarity Coefficient (DSC) serves as a widely recognized statistical metric for gauging the accuracy of image segmentation. This coefficient quantifies the similarity between two sets of data, typically represented as binary arrays, by comparing a predicted segmentation mask to the ground-truth mask. The DSC operates on a scale from one to zero, with one signifying a perfect match and zero indicating a complete mismatch. The utility of this metric lies in its ability to discern performance differences between classifiers, rendering it an invaluable instrument for evaluating segmentation algorithms. The DSC can be calculated as follows:

\begin{equation}
\label{DSC}
\operatorname{DSC}\left(m\left(x_{i}\right), y_{i}\right)=\frac{\left|2 *\left(m\left(x_{i}\right) \cap y_{i}\right)\right|}{\left|\left(m\left(x_{i}\right) \cap y_{i}\right)\right|+\left|m\left(x_{i}\right) \cup y_{i}\right|},
\end{equation}
where $\left|m\left(x_{i}\right) \cap y_{i}\right|$ and $\left|m\left(x_{i}\right) \cup y_{i}\right|$ refer to the area of overlap and area of union, respectively.
\section{Results and Discussion}

In this section, we present the results of our comprehensive experiments conducted on six datasets, employing eight prompting strategies, and utilizing three variations of the SAM. The performance of these models is compared to the current state-of-the-art (SOTA) methods, with certain zero-shot results of SAM surpassing established benchmarks. We subsequently engage in a qualitative discussion of the observed results, showcasing select challenging images to elucidate our findings. Finally, we provide a practical implementation guideline for physicians to effectively utilize the SAM, ensuring minimal interaction and delivering robust outcomes.

\begin{figure}[!htpb]
    \centering
    \includegraphics[width=.90\textwidth]{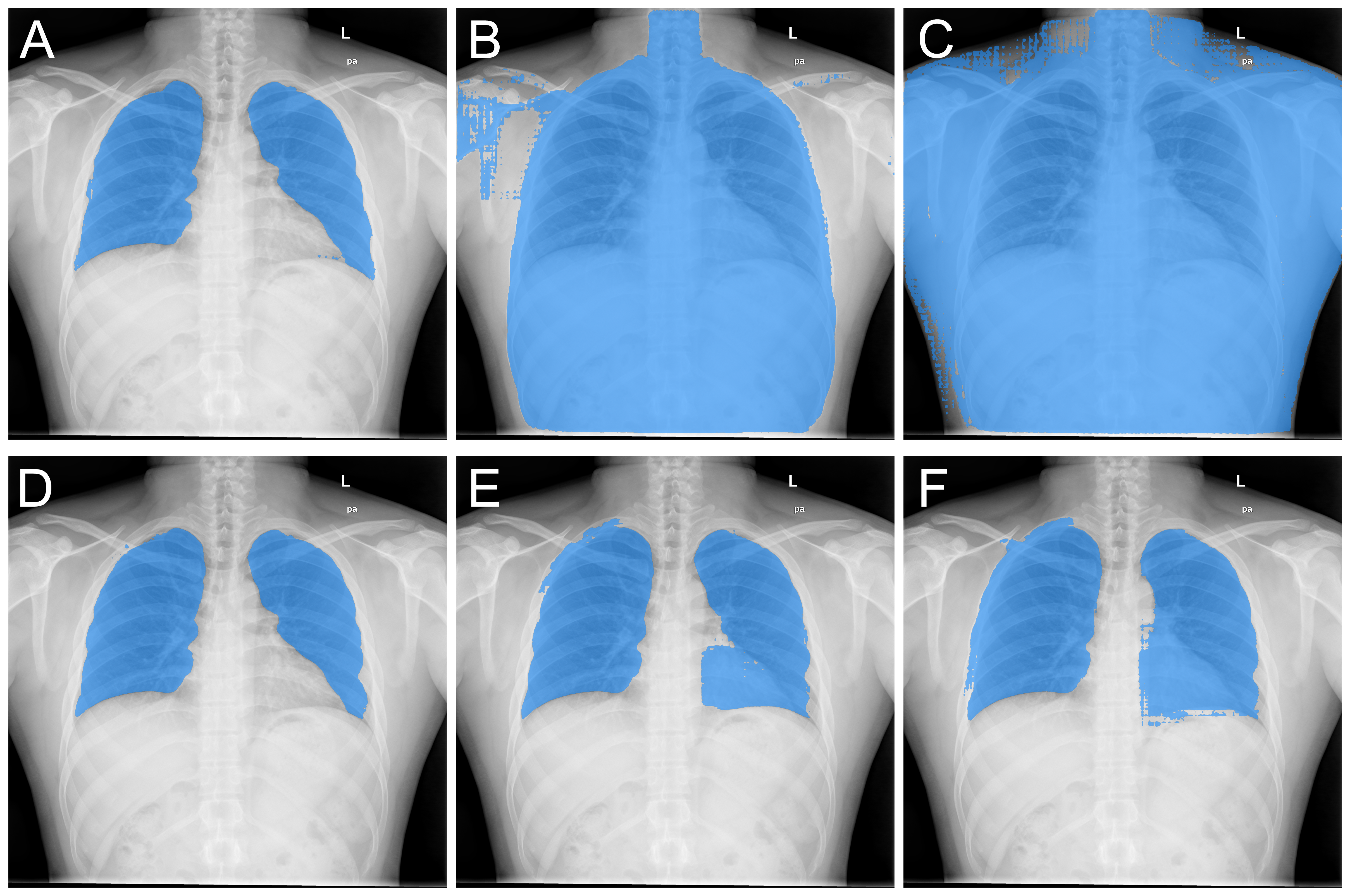}
    \caption{Three returning predictions from SAM using RP5 (A, B, C) and BBS10 (D, E, F) input methods for the CXR dataset. A physician may choose the one that best fits the corresponding region to be segmented.}
    \label{3images}
\end{figure}

Table~\ref{tab:results} shows the DSC of the predictions for ViT-H, the largest SAM model, with results for ViT-B and ViT-L shown in the Supplementary Material. The terms 1st, 2nd, and 3rd correspond to the three predictions generated by SAM, and the table presents the metrics when only one of these predictions is used consistently for all images. Fig~\ref{3images} showcases an example of these predictions for the Chest X-Ray (CXR) dataset, employing both the RP5 and BBS10 strategies. The RP5 method provides better differentiation between predictions, while the BBS10 approach demonstrates greater uniformity. This observation could potentially be attributed to the bounding box, which simultaneously indicates the target region for segmentation and the areas to be excluded (outside the box).

\begin{table}[!hp]
\caption{DSC of predictions for the ViT-H model for six datasets using the eight proposed prompt strategies considering the 1st, 2nd, and 3rd prediction.}
\label{tab:results}
\centering
\resizebox{\textwidth}{!}{%
\tiny
\begin{tabular}{@{}c@{\hspace{3pt}}l@{\hspace{3pt}}>{\columncolor{gray2}}S[table-format=1.3]@{\hspace{3pt}}>{\columncolor{gray5}}S[table-format=1.3]@{\hspace{3pt}}>{\columncolor{gray2}}S[table-format=1.3]@{\hspace{3pt}}>{\columncolor{gray5}}S[table-format=1.3]@{\hspace{3pt}}>{\columncolor{gray2}}S[table-format=1.3]@{\hspace{3pt}}>{\columncolor{gray5}}S[table-format=1.3]@{\hspace{3pt}}>{\columncolor{gray2}}S[table-format=1.3]@{\hspace{3pt}}>{\columncolor{gray5}}S[table-format=1.3]@{}}
\toprule
Dataset & Pred & \multicolumn{1}{>{\columncolor{white}}c}{CP} & \multicolumn{1}{>{\columncolor{white}}c}{RP} & \multicolumn{1}{>{\columncolor{white}}c}{RP3} & \multicolumn{1}{>{\columncolor{white}}c}{RP5} & \multicolumn{1}{>{\columncolor{white}}c}{BB} & \multicolumn{1}{>{\columncolor{white}}c}{BBS5} & \multicolumn{1}{>{\columncolor{white}}c}{BBS10} & \multicolumn{1}{>{\columncolor{white}}c}{BBS20} \\ \midrule
\multirow{3}{*}{ISIC}   & 1st & 0.538 & 0.531 & 0.762 & 0.774 & 0.745 & 0.737 & 0.715 & 0.603 \\
                        & 2nd & 0.718 & 0.677 & 0.769 & 0.788 & 0.845 & 0.842 & 0.833 & 0.789 \\
                        & 3rd & 0.375 & 0.363 & 0.390 & 0.483 & \textbf{0.872} & 0.868 & 0.860 & 0.816 \\ \midrule
\multirow{3}{*}{HAM}    & 1st & 0.544 & 0.527 & 0.752 & 0.765 & 0.732 & 0.724 & 0.700 & 0.589 \\
                        & 2nd & 0.729 & 0.686 & 0.768 & 0.785 & 0.838 & 0.835 & 0.824 & 0.778 \\
                        & 3rd & 0.420 & 0.406 & 0.443 & 0.541 & \textbf{0.865} & 0.861 & 0.851 & 0.809 \\ \midrule
\multirow{3}{*}{CXR}    & 1st & 0.904 & 0.863 & 0.923 & 0.927 & 0.936 & 0.934 & 0.911 & 0.686 \\
                        & 2nd & 0.758 & 0.727 & 0.766 & 0.828 & \textbf{0.942} & 0.939 & 0.929 & 0.826 \\
                        & 3rd & 0.471 & 0.469 & 0.482 & 0.514 & 0.935 & 0.930 & 0.913 & 0.803 \\ \midrule
\multirow{3}{*}{HJXR-F} & 1st & 0.876 & 0.822 & 0.941 & 0.948 & 0.924 & 0.908 & 0.848 & 0.618 \\
                        & 2nd & 0.743 & 0.767 & 0.767 & 0.776 & \textbf{0.962} & 0.958 & 0.904 & 0.746 \\
                        & 3rd & 0.517 & 0.543 & 0.548 & 0.599 & 0.949 & 0.945 & 0.905 & 0.723 \\ \midrule
\multirow{3}{*}{HJXR-I} & 1st & 0.211 & 0.742 & 0.808 & 0.828 & \textbf{0.875} & 0.866 & 0.734 & 0.624 \\
                        & 2nd & 0.393 & 0.479 & 0.449 & 0.491 & 0.855 & 0.849 & 0.790 & 0.620 \\
                        & 3rd & 0.294 & 0.295 & 0.316 & 0.384 & 0.800 & 0.796 & 0.758 & 0.629 \\ \midrule
\multirow{3}{*}{CVC}    & 1st & 0.716 & 0.763 & 0.861 & 0.880 & 0.889 & 0.881 & 0.835 & 0.702 \\
                        & 2nd & 0.554 & 0.544 & 0.642 & 0.754 & \textbf{0.926} & 0.924 & 0.916 & 0.844 \\
                        & 3rd & 0.232 & 0.224 & 0.224 & 0.245 & 0.924 & 0.922 & 0.918 & 0.868 \\ \midrule
\multirow{3}{*}{BUSI}   & 1st & 0.583 & 0.541 & 0.736 & 0.766 & 0.754 & 0.744 & 0.713 & 0.631 \\
                        & 2nd & 0.641 & 0.616 & 0.688 & 0.735 & 0.840 & 0.837 & 0.823 & 0.768 \\
                        & 3rd & 0.192 & 0.184 & 0.196 & 0.254 & \textbf{0.863} & 0.859 & 0.848 & 0.800 \\ \bottomrule
\end{tabular}%
}
\end{table}

\begin{figure}
%\begin{adjustwidth}{-2.25in}{0in}
    \centering
    \includegraphics[width=\textwidth]{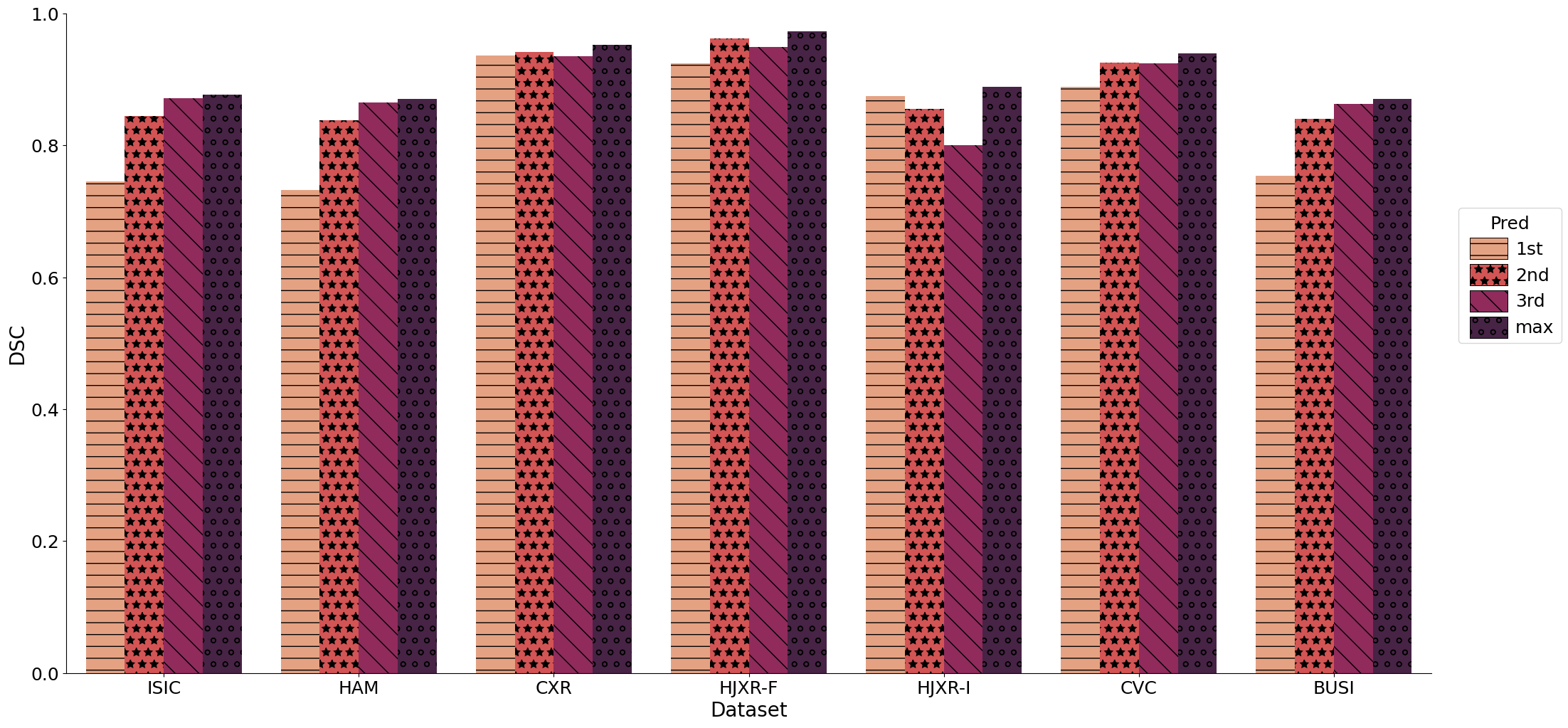}
    \caption{Comparison of using always the 1st, 2nd, or 3rd prediction versus choosing the best one per image (max) with the BB strategy for all datasets.}
    \label{graph}
%\end{adjustwidth}
\end{figure}

In a real-world clinical setting, a physician may opt to select the most suitable prediction. To simulate this decision-making process, we assessed the highest DSC per image, irrespective of being the 1st, 2nd, or 3rd prediction. The results are presented in Table~\ref{tab:max}. 
This approach led to a modest improvement of approximately 1\% compared to using only the 1st, 2nd or 3rd prediction in all images, as shown in Fig~\ref{graph}. Even though the overall enhancement is marginal, it holds significance for certain subjects and necessitates minimal input from the physician.

\begin{table}[!htpb]
\caption{DSC of predictions for all variations of SAM for six datasets using the eight proposed prompt strategies. For each set of predictions, only the one with the highest DSC was considered.}
\label{tab:max}
\centering
\resizebox{\textwidth}{!}{
\tiny
\begin{tabular}{@{}cl>{\columncolor{gray2}}l>{\columncolor{gray5}}l>{\columncolor{gray2}}l>{\columncolor{gray5}}l>{\columncolor{gray2}}l>{\columncolor{gray5}}l>{\columncolor{gray2}}l>{\columncolor{gray5}}l@{}}
\toprule
Dataset &
  \multicolumn{1}{c}{Model} &
  \multicolumn{1}{c}{CP} &
  \multicolumn{1}{c}{RP} &
  \multicolumn{1}{c}{RP3} &
  \multicolumn{1}{c}{RP5} &
  \multicolumn{1}{c}{BB} &
  \multicolumn{1}{c}{BBS5} &
  \multicolumn{1}{c}{BBS10} &
  \multicolumn{1}{c}{BBS20} \\ \midrule
\multirow{3}{*}{ISIC}   & ViT-H & 0.788 & 0.768 & 0.820 & 0.835 & {\ul 0.877}    & 0.874          & 0.866 & 0.829 \\
                        & ViT-L & 0.783 & 0.768 & 0.811 & 0.818 & {\ul 0.876}    & 0.872          & 0.864 & 0.819 \\
                        & ViT-B & 0.764 & 0.733 & 0.804 & 0.815 & \textbf{0.879} & 0.876          & 0.864 & 0.822 \\ \midrule
\multirow{3}{*}{HAM}    & ViT-H & 0.782 & 0.764 & 0.812 & 0.824 & {\ul 0.870}    & 0.866          & 0.857 & 0.820 \\
                        & ViT-L & 0.784 & 0.772 & 0.809 & 0.819 & {\ul 0.867}    & 0.864          & 0.854 & 0.809 \\
                        & ViT-B & 0.745 & 0.706 & 0.785 & 0.796 & \textbf{0.872} & 0.867          & 0.855 & 0.810 \\ \midrule
\multirow{3}{*}{CXR}    & ViT-H & 0.922 & 0.902 & 0.928 & 0.936 & {\ul 0.952}    & 0.950          & 0.942 & 0.862 \\
                        & ViT-L & 0.929 & 0.917 & 0.932 & 0.930 & \textbf{0.954} & 0.952          & 0.943 & 0.849 \\
                        & ViT-B & 0.915 & 0.893 & 0.930 & 0.935 & {\ul 0.948}    & 0.943          & 0.932 & 0.858 \\ \midrule
\multirow{3}{*}{HJXR-F} & ViT-H & 0.906 & 0.917 & 0.943 & 0.950 & \textbf{0.973} & \textbf{0.973} & 0.957 & 0.861 \\
                        & ViT-L & 0.910 & 0.916 & 0.939 & 0.948 & \textbf{0.973} & \textbf{0.973} & 0.956 & 0.880 \\
                        & ViT-B & 0.927 & 0.882 & 0.910 & 0.907 & {\ul 0.971}    & 0.969          & 0.950 & 0.870 \\ \midrule
\multirow{3}{*}{HJXR-I} & ViT-H & 0.483 & 0.786 & 0.808 & 0.828 & {\ul 0.889}    & 0.886          & 0.843 & 0.719 \\
                        & ViT-L & 0.478 & 0.841 & 0.865 & 0.860 & \textbf{0.894} & 0.889          & 0.839 & 0.726 \\
                        & ViT-B & 0.500 & 0.765 & 0.825 & 0.830 & {\ul 0.875}    & 0.870          & 0.838 & 0.696 \\ \midrule
\multirow{3}{*}{CVC}    & ViT-H & 0.838 & 0.854 & 0.884 & 0.898 & \textbf{0.940} & 0.938          & 0.934 & 0.889 \\
                        & ViT-L & 0.815 & 0.823 & 0.848 & 0.847 & {\ul 0.934}    & 0.931          & 0.920 & 0.869 \\
                        & ViT-B & 0.739 & 0.749 & 0.783 & 0.784 & {\ul 0.932}    & 0.930          & 0.921 & 0.851 \\ \midrule
\multirow{3}{*}{BUSI}   & ViT-H & 0.732 & 0.706 & 0.791 & 0.816 & {\ul 0.870}    & 0.868          & 0.855 & 0.813 \\
                        & ViT-L & 0.744 & 0.727 & 0.800 & 0.807 & {\ul 0.875}    & 0.872          & 0.865 & 0.810 \\
                        & ViT-B & 0.734 & 0.701 & 0.804 & 0.818 & \textbf{0.886} & 0.884          & 0.874 & 0.831 \\ \bottomrule
\end{tabular}%
}
\end{table}

The Bounding Box (BB) strategy consistently exhibited superior performance across all datasets, as illustrated in Table~\ref{tab:results} and Table~\ref{tab:max}. Even with variations of $5\%$ or $10$\% (BBS5, BBS10), this method outperforms all point prompt strategies, while BBS20 achieved results comparable to RP5. This observation underscores the robustness of the bounding box approach, even in the presence of minor inaccuracies while delineating the desired segmentation region.

Regarding point prompt methods (CP, RP, RP3, RP5), an increased number of input points correspond to enhanced model performance. However, these techniques could not outperform the BB, BBS5, and BBS10 strategies. Moreover, RP5 requires greater manual intervention, rendering it more labor-intensive compared to employing a bounding box.

Our experiments do not incorporate additional prompt points that can be introduced post-prediction to refine the segmentation. This fine-tuning process can be applied to both encompass regions excluded from the prediction and eliminate regions that should not be part of the segmentation. As a result, physicians can achieve even more precise segmentation masks with minimal additional effort.

We also highlight that the ViT-B model attained performance levels comparable to the larger variants of SAM, occasionally even surpassing them. Furthermore, owing to its modest GPU memory requirements, it can be readily utilized with cost-effective hardware, making SAM's application in medical imaging highly accessible without a significant cost.

\subsection{Comparison with state-of-the-art (SOTA) segmentation models}

We employ the intermediate-sized SAM (ViT-L) for a comparative analysis with the current state-of-the-art (SOTA) models. Table~\ref{tab:compar} presents a performance comparison of SAM with the BB5 strategy (emulating a physician annotating with minimal error, followed by selecting the most accurate among three predictions) against SOTA models employed on each dataset. Notably, no baseline models were found for evaluation in the HJXR dataset.

\begin{table}[!htpb]
    \caption{Comparison of the results of the BBS5 strategy using the ViT-L model with the current state-of-the-art DL models.}
    \label{tab:compar}
    \centering
    \renewcommand{\arraystretch}{1.5}
    \scriptsize
    \begin{tabularx}{\textwidth}{cXl}
    \toprule
    \textbf{Dataset}       & \textbf{Model}            & \textbf{DSC} \\ \midrule
    \multirow{2}{*}{ISIC}  & Rema-net \citep{yang2023rema}            & \textbf{0.944} \\
                           & SAM ViT-L BBS5                     & 0.872 \\ \midrule
    \multirow{2}{*}{HAM}   & Rema-net \citep{yang2023rema}            & \textbf{0.936} \\
                           & SAM ViT-L BBS5                       & 0.864 \\ \midrule
    \multirow{3}{*}{CXR}   & Attention U-Net \citep{kim2021automatic} & \textbf{0.982} \\
                           & ReSE-Net \citep{agrawalrese}             & 0.976 \\
                           & SAM ViT-L BBS5                       & 0.952 \\ \midrule
    \multirow{2}{*}{CVC}   & FSA-Net~\citep{zhan2023fsa}              & \textbf{0.947} \\
                           & SAM ViT-L BBS5                       & 0.931 \\ \midrule
    \multirow{2}{*}{BUSI}  & PODDA, A. et al \citep{podda2022fully}   & 0.826 \\
                           & SAM ViT-L BBS5                       & \textbf{0.872} \\ \midrule
    \multirow{1}{*}{HJXR-F}& SAM ViT-L BBS5                       & \textbf{0.973} \\ \midrule
    \multirow{1}{*}{HJXR-I}& SAM ViT-L BBS5                       & \textbf{0.889} \\ \bottomrule
    \end{tabularx}
\end{table}

SAM achieved very strong results for a zero-shot (no training/fine-tuning) approach in comparison to the SOTA. In the BUSI dataset, SAM surpassed the SOTA by approximately 5\%, sustaining its superior performance even when employing the BBS20 strategy, which accommodates a substantial margin of error in image annotation. In the CVC dataset, SAM's performance was marginally lower (less than 2\%), while in the CXR dataset, the gap was a mere 3\%.

Although no directly comparable studies exist, SAM exhibited a very high DSC ($0.973$) for femur segmentation. The segmentation of ilios is a more intricate task due to reduced contrast with adjacent regions. Taking that into account, the results for ilios segmentation can also be considered quite strong.

For the ISIC and HAM datasets, SAM was outperformed by $\approx 7$\%. But here we need to take into account the unique characteristics of these datasets, and a more nuanced analysis is presented in the next section. Moreover, the substantial volume of available data (over $10,000$ images) renders the training of task-specific deep learning (DL) models more viable for those tasks. In contrast, with smaller datasets like BUSI, training an end-to-end DL model becomes strenuous due to the scarcity of data. In such scenarios, employing a model like SAM proves to be the best option, as it benefits from exposure to an extensive range of data across various domains.

\subsection{Qualitative Analysis}
\label{qualitative}

\begin{figure}[!htpb]
    \centering
    \includegraphics[width=0.6\columnwidth]{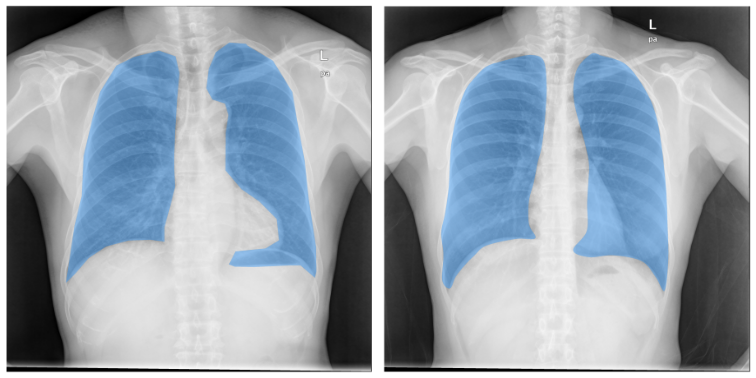}
    \caption{Example of inconsistencies within the ground-truth region in the CXR dataset.}
    \label{mask_crx}
\end{figure}

The analysis of medical images presents a unique set of challenges due to the complex and diverse nature of datasets. For instance, the CXR dataset, which consists of chest X-rays and their corresponding segmentation masks, contains inconsistencies in the segmented regions, as depicted in Fig~\ref{mask_crx}. Some ground-truth masks include the heart while others exclude it. Still, SAM can rapidly rectify these discrepancies by allowing users to select the most appropriate prediction, as demonstrated in Fig~\ref{3images}, or by refining input points to include or exclude specific regions as needed.

The standard DICOM format for X-ray images typically features a $12$ or $16$ bit depth, enabling physicians to manipulate the window/level settings for enhanced visualization of tissues and organs. We postulate that optimizing the window/level parameters during conversion to JPEG or PNG formats could improve tissue delineation and subsequently enhance SAM's performance for this imaging modality. Nevertheless, we did not assess this approach, given that the CXR dataset is provided in PNG format, and the HJXR dataset was normalized and converted to PNG using its maximum and minimum values.

\begin{figure}[!htpb]
    \centering
    \includegraphics[width=.70\textwidth]{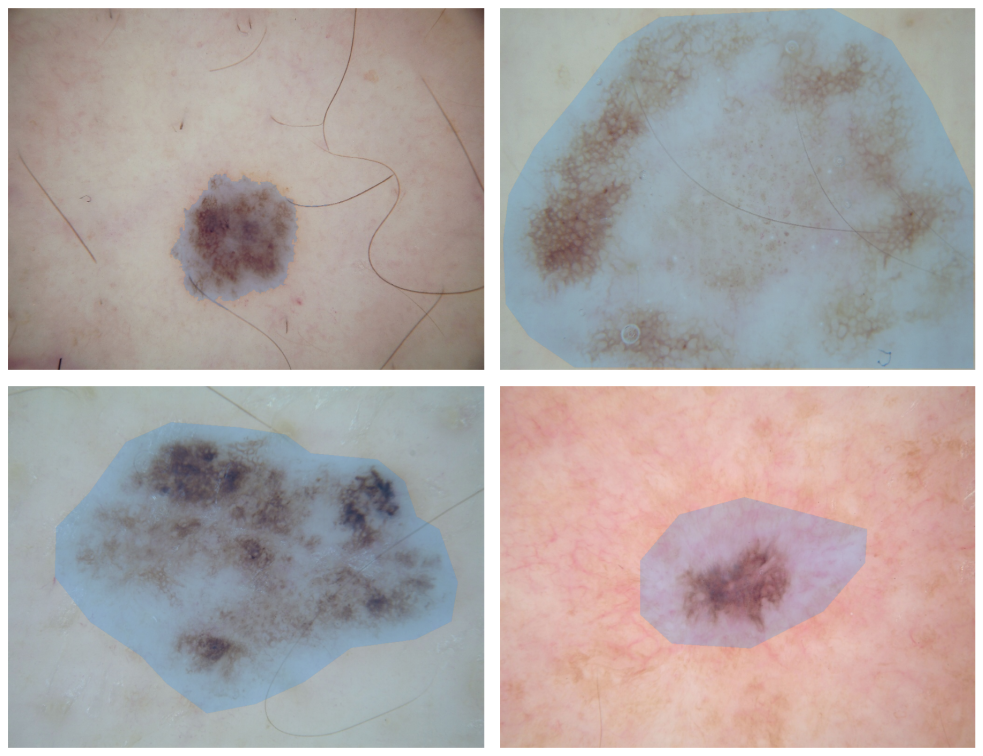}
    \caption{Example of inconsistencies in the ground-truth region in the ISIC dataset.}
    \label{mask_isic}
\end{figure}

For the ISIC dataset, which comprises skin lesion images, we identified numerous instances of inaccurate ground-truth mask annotations, as illustrated in Fig~\ref{mask_isic}. These inaccuracies impacted the DSC results, as the masks generated by SAM appear to exhibit higher precision compared to the original masks. Moreover, the presence of body hair in the ISIC and HAM datasets significantly influences the segmentation process, particularly when employing point prompt strategies. For example, a hair intersecting a lesion may erroneously indicate two distinct regions instead of one. To address this issue, bounding box strategies can be implemented to provide sufficient information to the model. However, SAM's exclusion of hair from the segmentation negatively affects its performance. Additionally, the skin lesion datasets present challenges due to indistinct lesion boundaries, rendering accurate segmentation of skin lesions a challenging task.

Ultrasound images pose considerable difficulties for DL models, attributable to their inhomogeneous intensities and low signal-to-noise ratio, which hinder the accurate delineation of breast tumors in datasets such as BUSI. Furthermore, the absorption and reflection of ultrasound can give rise to artifacts in the image, exacerbating the segmentation task even for well-optimized models. Despite these obstacles, SAM achieved strong results in this dataset. However, it encountered challenges in accurately segmenting the boundaries of breast tumors due to the inherent blurred edges in ultrasound images.

\subsection{Guidelines}

In light of our empirical findings, we propose a robust and pragmatic framework for utilizing the Segment Anything Model (SAM) in the realm of medical imaging tasks. This methodology empowers physicians to capitalize on the capabilities of SAM to attain precise segmentation outcomes, while preserving their autonomy in overseeing the process. Our recommendation is to employ the largest SAM variation that is feasible given the constraints of the available hardware; nonetheless, any of the three model variants may be utilized.

\begin{enumerate}

\item \textbf{Initiate with a bounding box prompt}: our experimental results consistently indicate that among various prompting strategies, the bounding box technique exhibits superior performance, even in the presence of minor variations. Thus, we advocate that physicians start the segmentation procedure by supplying a bounding box prompt encompassing the region of interest.

\item \textbf{Evaluate the generated predictions}: SAM generates a triplet of segmentation masks in response to an input image and a bounding box, each signifying a distinct interpretation of the intended region's dimensions. Physicians are advised to visually scrutinize and juxtapose the three produced masks against the source image. If there is a suitable prediction, select it. If none of the predictions correctly segment the intended region, proceed to the next step.

\item \textbf{Refine the segmentation employing point prompts}: in cases where none of the initial predictions adequately segment the intended region, assess the best prediction and identify the areas it incorrectly captures or omits in the segmentation. Utilize input points to include (label $1$) or exclude (label $0$) these areas. SAM will generate three new predictions. Repeat the process of refining the segmentation using point prompts until an adequate segmentation is achieved.

\end{enumerate}

\begin{figure}[h]
    \centering
    \includegraphics[width=\textwidth]{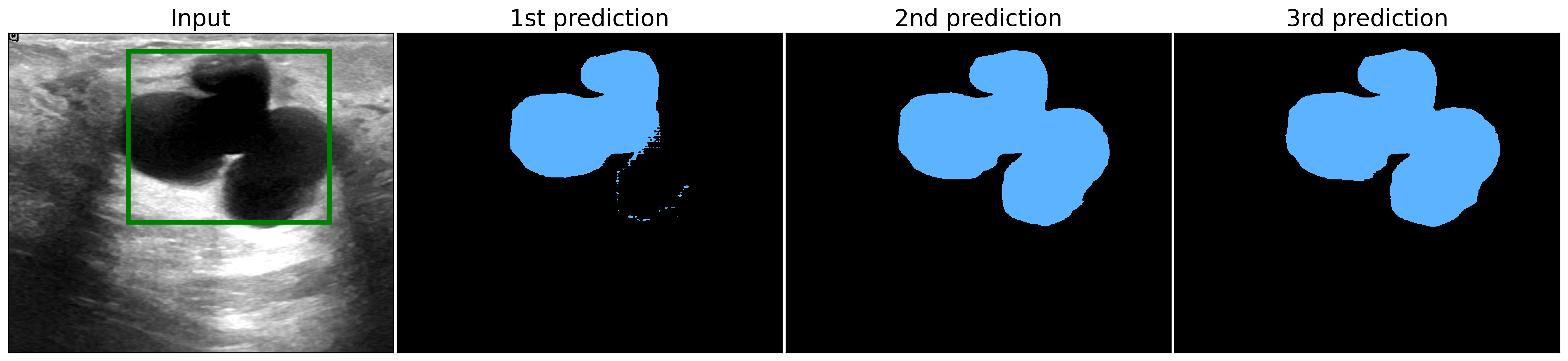}
    \caption{Image from the BUSI dataset with bounding box input accompanied by SAM's predictions. Both the 2nd and 3rd predictions exhibit accurate segmentation of the intended region.}
    \label{guideline0}
\end{figure}

\begin{figure}[h]
    \centering
    \includegraphics[width=\textwidth]{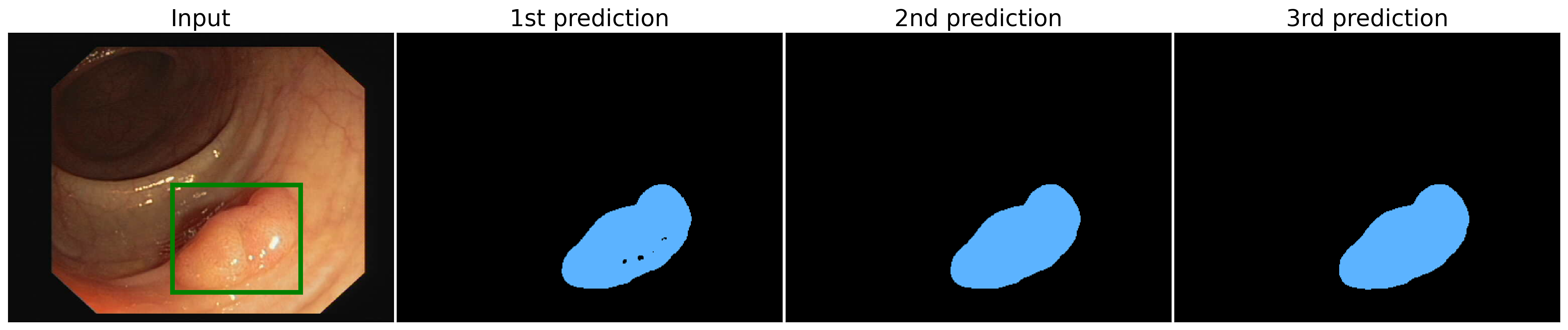}
    \caption{Image from the CVC dataset with bounding box input accompanied by SAM's predictions. Both the 2nd and 3rd predictions exhibit accurate segmentation of the intended region.}
    \label{guideline1}
\end{figure}

Fig.~\ref{guideline0} and Fig.~\ref{guideline1} demonstrate the application of our proposed framework on images from the BUSI and CVC datasets, including the bounding box prompt and subsequent predictions. Since the intended regions were accurately segmented, the physician merely has to select them.

\begin{figure}[h]
    \centering
    \includegraphics[width=\textwidth]{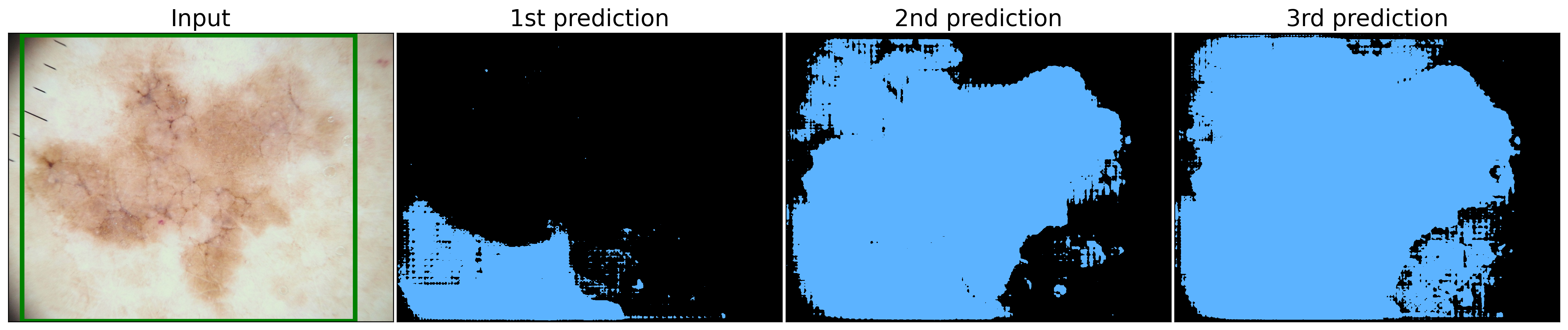}
    \caption{Image from the ISIC dataset with bounding box input accompanied by SAM's predictions. None of them are adequate and require further prompt points.}
    \label{guideline2}
\end{figure}

\begin{figure}[h]
    \centering
    \includegraphics[width=\textwidth]{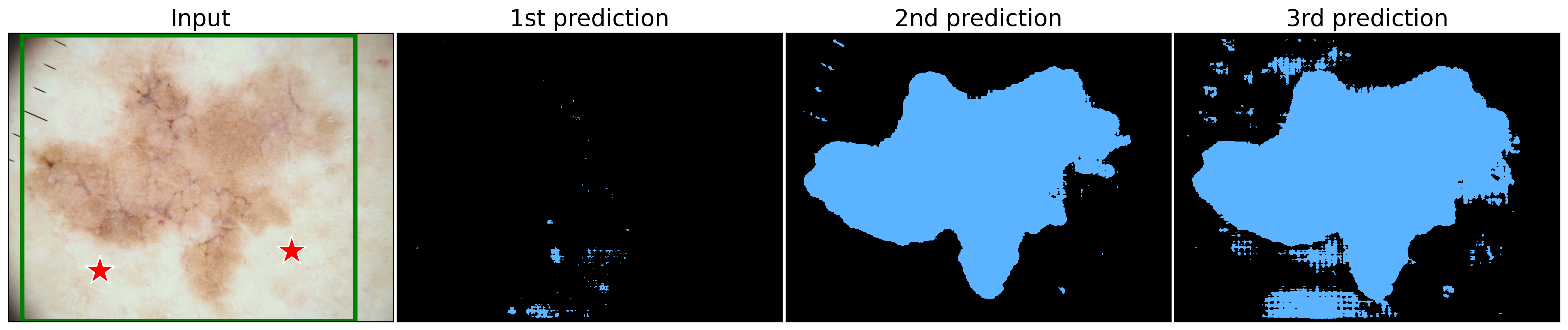}
    \caption{Image from the ISIC dataset with bounding box and point inputs accompanied by SAM's predictions. The point prompts guide the model to remove these areas. The 2nd prediction reached an adequate segmentation.}
    \label{guideline3}
\end{figure}

Fig.~\ref{guideline2} presents the application of our framework on an image from the ISIC dataset, followed by the bounding box prompt and predictions. This represents a more intricate scenario, as discussed earlier. None of the predictions provided satisfactory results; therefore, the physician must evaluate the best one (2nd) and incorporate point prompts to guide the model. Fig \ref{guideline3} displays the original bounding box input in conjunction with the point prompts and the generated predictions. A significant improvement in segmentation is observable in the 2nd prediction due to the additional point prompts.

This methodology ensures that the model's output coheres with the physician's expertise, culminating in accurate and dependable segmentation results across diverse clinical applications and imaging modalities.\footnote{A demo of this framework is available at \url{https://github.com/Malta-Lab/SAM-zero-shot-in-Medical-Imaging}.}
\section{Conclusion and Future Work}

In this study, we thoroughly evaluated the zero-shot performance of SAM by employing eight distinct prompting strategies across six datasets from four different 2D medical image modalities. Our comprehensive analysis shed light on the advantages and limitations of these strategies in various scenarios for the three SAM ViT sizes. Remarkably, SAM demonstrated exceptional performance as a zero-shot approach, achieving competitive results in comparison to the state-of-the-art segmentation methods specifically designed or fine-tuned for a particular modality of medical imaging. Notably, SAM outperformed the current best performance on the BUSI dataset by a substantial margin. Taken together, our findings underscore the immense potential of SAM as a powerful tool for low-effort medical image segmentation.

Drawing upon our results, we propose pragmatic guidelines that facilitate easy implementation, necessitate minimal user interaction, and yield robust outcomes in medical imaging segmentation with SAM. By incorporating the bounding box method and refining the segmentation using point prompts, medical practitioners can effectively harness SAM's potential to attain accurate results while maintaining control over the segmentation process. Furthermore, given the comparable performance of the three SAM sizes, practitioners can choose any of them based on their hardware resource constraints.

The segmentation results generated by SAM have the potential to exceed the most stringent quality standards with minimal involvement from physicians. Our findings highlight concerns regarding the quality of some manually-annotated ground truth masks, as SAM outcomes appear to delineate the region of interest more accurately in certain instances. This observation holds particular significance for labeling new datasets, as it substantially reduces the time and effort required for this laborious and tedious task. Consequently, SAM-generated segmentation masks offer immense promise for streamlining data annotation processes and enhancing workflow efficiency in the area of medical image analysis.

Future research endeavors could focus on further augmenting SAM's capabilities in this domain, achieving even higher performance while preserving SAM's extensive refinement options. Additionally, investigating the potential of adapting SAM for 3D medical imaging represents a valuable research direction, as it would extend the model's applicability to a broader spectrum of medical imaging tasks.

\bibliography{myref}
\newpage
\section{Supplementary Material}
\label{supllementary}

\begin{longtable}[c]{@{}cccllllllll@{}}
\toprule
Dataset & Model & Pred & \multicolumn{1}{c}{CP} & \multicolumn{1}{c}{RP} & \multicolumn{1}{c}{RP3} & \multicolumn{1}{c}{RP5} & \multicolumn{1}{c}{BB} & \multicolumn{1}{c}{BBS5} & \multicolumn{1}{c}{BBS10} & \multicolumn{1}{c}{BBS20} \\* \midrule
\endhead
\bottomrule
\endfoot
\endlastfoot
\multirow{9}{*}{ISIC} & \multirow{3}{*}{ViT-H} & 1st & 0.538 & 0.531 & 0.762 & 0.774 & 0.745 & 0.737 & 0.715 & 0.603 \\
 &  & 2nd & 0.718 & 0.677 & 0.769 & 0.788 & 0.845 & 0.842 & 0.833 & 0.789 \\
 &  & 3rd & 0.375 & 0.363 & 0.390 & 0.483 & \textbf{0.872} & 0.868 & 0.860 & 0.816 \\* \cmidrule(l){2-11} 
 & \multirow{3}{*}{ViT-L} & 1st & 0.704 & 0.665 & 0.703 & 0.700 & \textbf{0.864} & 0.861 & 0.852 & 0.805 \\
 &  & 2nd & 0.518 & 0.521 & 0.768 & 0.794 & 0.763 & 0.757 & 0.733 & 0.623 \\
 &  & 3rd & 0.382 & 0.366 & 0.358 & 0.362 & 0.841 & 0.836 & 0.819 & 0.730 \\* \cmidrule(l){2-11} 
 & \multirow{3}{*}{ViT-B} & 1st & 0.366 & 0.355 & 0.354 & 0.375 & \textbf{0.870} & 0.866 & 0.855 & 0.810 \\
 &  & 2nd & 0.665 & 0.618 & 0.692 & 0.695 & 0.825 & 0.823 & 0.807 & 0.751 \\
 &  & 3rd & 0.504 & 0.490 & 0.766 & 0.790 & 0.640 & 0.631 & 0.601 & 0.496 \\* \midrule
\multirow{9}{*}{HAM} & ViT-H & 1st & 0.544 & 0.527 & 0.752 & 0.765 & 0.732 & 0.724 & 0.700 & 0.589 \\
 &  & 2nd & 0.729 & 0.686 & 0.768 & 0.785 & 0.838 & 0.835 & 0.824 & 0.778 \\
 &  & 3rd & 0.420 & 0.406 & 0.443 & 0.541 & \textbf{0.865} & 0.861 & 0.851 & 0.809 \\* \cmidrule(l){2-11} 
 & \multirow{3}{*}{ViT-L} & 1st & 0.731 & 0.689 & 0.723 & 0.721 & \textbf{0.859} & 0.856 & 0.846 & 0.799 \\
 &  & 2nd & 0.522 & 0.518 & 0.764 & 0.793 & 0.766 & 0.761 & 0.740 & 0.626 \\
 &  & 3rd & 0.435 & 0.413 & 0.406 & 0.408 & 0.830 & 0.824 & 0.805 & 0.699 \\* \cmidrule(l){2-11} 
 & \multirow{3}{*}{ViT-B} & 1st & 0.414 & 0.403 & 0.401 & 0.425 & \textbf{0.863} & 0.859 & 0.846 & 0.799 \\
 &  & 2nd & 0.659 & 0.607 & 0.681 & 0.683 & 0.810 & 0.807 & 0.795 & 0.740 \\
 &  & 3rd & 0.478 & 0.431 & 0.749 & 0.772 & 0.619 & 0.610 & 0.578 & 0.466 \\* \midrule
\multirow{9}{*}{CXR} & \multirow{3}{*}{ViT-H} & 1st & 0.904 & 0.863 & 0.923 & 0.927 & 0.936 & 0.934 & 0.911 & 0.686 \\
 &  & 2nd & 0.758 & 0.727 & 0.766 & 0.828 & \textbf{0.942} & 0.939 & 0.929 & 0.826 \\
 &  & 3rd & 0.471 & 0.469 & 0.482 & 0.514 & 0.935 & 0.930 & 0.913 & 0.803 \\* \cmidrule(l){2-11} 
 & \multirow{3}{*}{ViT-L} & 1st & 0.834 & 0.814 & 0.786 & 0.776 & 0.932 & 0.929 & 0.916 & 0.805 \\
 &  & 2nd & 0.915 & 0.870 & 0.930 & 0.929 & 0.940 & 0.936 & 0.906 & 0.660 \\
 &  & 3rd & 0.472 & 0.471 & 0.468 & 0.474 & \textbf{0.945} & 0.942 & 0.928 & 0.758 \\* \cmidrule(l){2-11} 
 & \multirow{3}{*}{ViT-B} & 1st & 0.459 & 0.459 & 0.467 & 0.497 & 0.916 & 0.910 & 0.894 & 0.817 \\
 &  & 2nd & 0.804 & 0.782 & 0.786 & 0.803 & \textbf{0.937} & 0.933 & 0.921 & 0.813 \\
 &  & 3rd & 0.882 & 0.813 & 0.928 & 0.932 & 0.916 & 0.898 & 0.818 & 0.524 \\* \midrule
\multirow{9}{*}{HJXR-F} & \multirow{3}{*}{ViT-H} & 1st & 0.876 & 0.822 & 0.941 & 0.948 & 0.924 & 0.908 & 0.848 & 0.618 \\
 &  & 2nd & 0.743 & 0.767 & 0.767 & 0.776 & \textbf{0.962} & 0.958 & 0.904 & 0.746 \\
 &  & 3rd & 0.517 & 0.543 & 0.548 & 0.599 & 0.949 & 0.945 & 0.905 & 0.723 \\* \cmidrule(l){2-11} 
 & \multirow{3}{*}{ViT-L} & 1st & 0.773 & 0.800 & 0.788 & 0.791 & \textbf{0.972} & 0.969 & 0.951 & 0.843 \\
 &  & 2nd & 0.874 & 0.804 & 0.927 & 0.944 & 0.925 & 0.922 & 0.844 & 0.685 \\
 &  & 3rd & 0.516 & 0.540 & 0.540 & 0.619 & 0.961 & 0.944 & 0.818 & 0.448 \\* \cmidrule(l){2-11} 
 & \multirow{3}{*}{ViT-B} & 1st & 0.466 & 0.486 & 0.481 & 0.489 & 0.924 & 0.915 & 0.888 & 0.788 \\
 &  & 2nd & 0.733 & 0.775 & 0.742 & 0.727 & \textbf{0.958} & 0.954 & 0.926 & 0.771 \\
 &  & 3rd & 0.911 & 0.774 & 0.909 & 0.907 & 0.899 & 0.876 & 0.735 & 0.490 \\* \midrule
\multirow{9}{*}{HJXR-I} & \multirow{3}{*}{ViT-H} & 1st & 0.211 & 0.742 & 0.808 & 0.828 & \textbf{0.875} & 0.866 & 0.734 & 0.624 \\
 &  & 2nd & 0.393 & 0.479 & 0.449 & 0.491 & 0.855 & 0.849 & 0.790 & 0.620 \\
 &  & 3rd & 0.294 & 0.295 & 0.316 & 0.384 & 0.800 & 0.796 & 0.758 & 0.629 \\* \cmidrule(l){2-11} 
 & \multirow{3}{*}{ViT-L} & 1st & 0.363 & 0.540 & 0.448 & 0.451 & 0.824 & 0.817 & 0.748 & 0.594 \\
 &  & 2nd & 0.165 & 0.758 & 0.864 & 0.860 & \textbf{0.887} & 0.877 & 0.762 & 0.555 \\
 &  & 3rd & 0.301 & 0.306 & 0.292 & 0.330 & 0.862 & 0.841 & 0.733 & 0.580 \\* \cmidrule(l){2-11} 
 & \multirow{3}{*}{ViT-B} & 1st & 0.259 & 0.303 & 0.328 & 0.368 & 0.772 & 0.767 & 0.734 & 0.591 \\
 &  & 2nd & 0.403 & 0.502 & 0.467 & 0.478 & \textbf{0.849} & 0.843 & 0.802 & 0.615 \\
 &  & 3rd & 0.314 & 0.717 & 0.823 & 0.830 & 0.838 & 0.838 & 0.779 & 0.622 \\* \midrule
\multirow{9}{*}{CVC} & \multirow{3}{*}{ViT-H} & 1st & 0.716 & 0.763 & 0.861 & 0.880 & 0.889 & 0.881 & 0.835 & 0.702 \\
 &  & 2nd & 0.554 & 0.544 & 0.642 & 0.754 & \textbf{0.926} & 0.924 & 0.916 & 0.844 \\
 &  & 3rd & 0.232 & 0.224 & 0.224 & 0.245 & 0.924 & 0.922 & 0.918 & 0.868 \\* \cmidrule(l){2-11} 
 & \multirow{3}{*}{ViT-L} & 1st & 0.498 & 0.482 & 0.508 & 0.522 & \textbf{0.920} & 0.918 & 0.906 & 0.853 \\
 &  & 2nd & 0.702 & 0.728 & 0.836 & 0.841 & 0.873 & 0.867 & 0.818 & 0.672 \\
 &  & 3rd & 0.229 & 0.222 & 0.217 & 0.223 & 0.909 & 0.904 & 0.870 & 0.773 \\* \cmidrule(l){2-11} 
 & \multirow{3}{*}{ViT-B} & 1st & 0.234 & 0.225 & 0.222 & 0.226 & \textbf{0.920} & 0.916 & 0.906 & 0.833 \\
 &  & 2nd & 0.447 & 0.440 & 0.495 & 0.510 & 0.907 & 0.906 & 0.892 & 0.796 \\
 &  & 3rd & 0.644 & 0.688 & 0.778 & 0.783 & 0.821 & 0.810 & 0.758 & 0.585 \\* \midrule
\multirow{9}{*}{BUSI} & \multirow{3}{*}{ViT-H} & 1st & 0.583 & 0.541 & 0.736 & 0.766 & 0.754 & 0.744 & 0.713 & 0.631 \\
 &  & 2nd & 0.641 & 0.616 & 0.688 & 0.735 & 0.840 & 0.837 & 0.823 & 0.768 \\
 &  & 3rd & 0.192 & 0.184 & 0.196 & 0.254 & \textbf{0.863} & 0.859 & 0.848 & 0.800 \\* \cmidrule(l){2-11} 
 & \multirow{3}{*}{ViT-L} & 1st & 0.656 & 0.649 & 0.674 & 0.663 & \textbf{0.866} & 0.862 & 0.855 & 0.794 \\
 &  & 2nd & 0.567 & 0.536 & 0.748 & 0.779 & 0.782 & 0.777 & 0.754 & 0.649 \\
 &  & 3rd & 0.228 & 0.205 & 0.202 & 0.252 & 0.849 & 0.847 & 0.830 & 0.741 \\* \cmidrule(l){2-11} 
 & \multirow{3}{*}{ViT-B} & 1st & 0.202 & 0.192 & 0.181 & 0.213 & \textbf{0.884} & 0.881 & 0.869 & 0.823 \\
 &  & 2nd & 0.634 & 0.604 & 0.682 & 0.691 & 0.832 & 0.830 & 0.818 & 0.766 \\
 &  & 3rd & 0.562 & 0.522 & 0.773 & 0.797 & 0.725 & 0.722 & 0.689 & 0.582 \\* \bottomrule
\caption{DSC of predictions for six datasets using the eight proposed prompt strategies for the three SAM ViT sizes.}
\label{tab:all}
\end{longtable}

\bibliographystyle{unsrtnat}
%\bibliographystyle{splncs04}
%\bibliography{mybibliography}
%

\end{document}